%% file: main.tex
\documentclass[10pt,twocolumn,letterpaper]{article}

\usepackage{cvpr}              

\usepackage{graphicx}
\usepackage{amsmath}
\usepackage{amssymb}
\usepackage{booktabs}
\usepackage{makecell}

\usepackage[linesnumbered,ruled,lined]{algorithm2e}

\SetCommentSty{mycommfont}
\usepackage{colortbl}
\usepackage{multirow}
\usepackage{graphicx}
\usepackage[normalem]{ulem}
\useunder{\uline}{\ul}{}

\usepackage{color, colortbl}
\definecolor{MozhganGray}{rgb}{0.95, 0.95, 0.95}

\newcommand\smallx[1]{
  \mathchoice
    {{\scriptstyle X}}
    {{\scriptstyle X}}
    {{\scriptscriptstyle X}}
    {\scalebox{.7}{$\scriptscriptstyle X$}}
  }

\newcommand\smally[1]{
  \mathchoice
    {{\scriptstyle Y}}
    {{\scriptstyle Y}}
    {{\scriptscriptstyle Y}}
    {\scalebox{.7}{$\scriptscriptstyle Y$}}
  }

  \newcommand\smalls[1]{
  \mathchoice
    {{\scriptstyle S}}
    {{\scriptstyle S}}
    {{\scriptscriptstyle S}}
    {\scalebox{.7}{$\scriptscriptstyle S$}}
  }

  \newcommand\smallyy[1]{
  \mathchoice
    {{\scriptstyle \mathcal{Y}}}
    {{\scriptstyle \mathcal{Y}}}
    {{\scriptscriptstyle \mathcal{Y}}}
    {\scalebox{.7}{$\scriptscriptstyle \mathcal{Y}$}}
  }

\input{preamble}

\definecolor{cvprblue}{rgb}{0.21,0.49,0.74}
\usepackage[pagebackref,breaklinks,colorlinks,citecolor=cvprblue]{hyperref}

\def\paperID{3269} 
\def\confName{CVPR}
\def\confYear{2024}

\title{Adversarial Backdoor Attack by Naturalistic Data Poisoning on Trajectory Prediction in Autonomous Driving}

\author{%
Mozhgan Pourkeshavarz$^{1}$ \quad Mohammad Sabokrou$^{2}$ \quad Amir Rasouli$^1$  \\
$^1$Noah’s Ark Lab, Huawei \quad  $^2$Okinawa Institute of Science and Technology (OIST) \\
\texttt{firstname.lastname@huawei.com} \\
\texttt{mohammad.sabokrou@oist.jp}
}

\begin{document}
\maketitle
\input{sec/0_abstract}    
\input{sec/1_intro}
\input{sec/2_related_work}
\input{sec/3_method}
\input{sec/4_experiments}
\input{sec/5_conclusion}

{
    \small
    \bibliographystyle{ieeenat_fullname}
    \bibliography{main}
}


\end{document}

%% file: preamble.tex
\usepackage[dvipsnames]{xcolor}


%% file: sec/0_abstract.tex
\begin{abstract}
In autonomous driving, behavior prediction is fundamental for safe motion planning, hence the security and robustness of prediction models against adversarial attacks are of paramount importance. We propose a novel adversarial backdoor attack against trajectory prediction models as a means of studying their potential vulnerabilities. Our attack affects the victim at training time via naturalistic, hence stealthy, poisoned samples crafted using a novel two-step approach. First, the triggers are crafted by perturbing the trajectory of attacking vehicle and then disguised by transforming the scene using a bi-level optimization technique. The proposed attack does not depend on a particular model architecture and operates in a black-box manner, thus can be effective without any knowledge of the victim model. We conduct extensive empirical studies using state-of-the-art prediction models on two benchmark datasets using metrics customized for trajectory prediction. We show that the proposed attack is highly effective, as it can significantly hinder the performance of prediction models, unnoticeable by the victims, and efficient as it forces the victim to generate malicious behavior even under constrained conditions. Via ablative studies, we analyze the impact of different attack design choices followed by an evaluation of existing defence mechanisms against the proposed attack.
\end{abstract}
\vspace{-5mm}

%% file: sec/1_intro.tex
\section{Introduction}
 \begin{figure}[t]
	\centering
	\includegraphics[width=1\columnwidth]{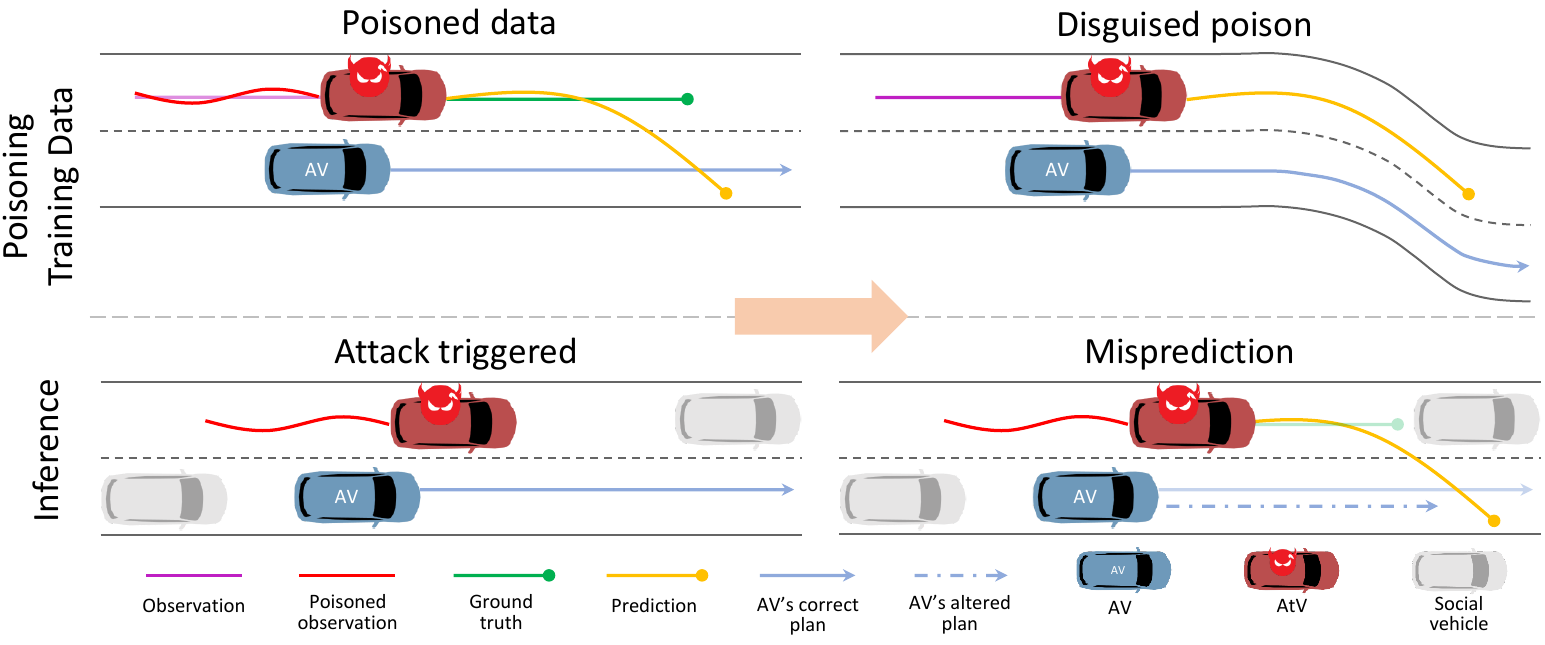}
\vspace{-6mm}
	\caption {Illustration of the proposed adversarial backdoor attack. Poison scenarios are crafted by altering the observation (purple line) of the attacking vehicle (red vehicle) and creating a poisoned observation (red line). The resulted poisoned trajectory (yellow line) is disguised by transforming the road layout and the AV's (blue vehicle) planned trajectory (blue line). The samples are then injected in the training data of the victim. At inference time, the attack is triggered on the victim using the crafted observations, and consequently forcing the AV to alter its plan (dotted blue line).}
	\label{fig:first_image}
	\vspace{-6mm}
\end{figure}

Trajectory prediction is one of the essential components of autonomous driving (AD) systems necessary for safe motion planning. Modern prediction models are designed based on deep neural networks (DNNs) \cite{zhou2022hivt, zhao2021tnt,deo2022multimodal,park2023leveraging,girgislatent2022, song2022learning} achieving promising performance on the existing AD benchmarks \cite{chang2019argoverse}. Meanwhile, with the widespread deployment of DNNs in real-world safety critical applications, such as AD, there is a growing concern about the security of these systems \cite{yu2023backdoor,yu2022towards, yi2021benchmarking}. 

There is a large body of literature on adversarial attacks and their impact on deep networks \cite{chou2023backdoor}. In the domain of trajectory prediction for AD,  a handful of attacks have been proposed \cite{zhang2022adversarial,bahari2022vehicle,cao2022advdo,Zheng_2023_WACV} aiming to alter the performance of the prediction models by introducing various perturbations to the dynamics of the agents surrounding autonomous vehicles or the scene. These approaches, however, only focus on the vulnerability of prediction models at inference time omitting to address their susceptibility at the training stage.   

In computer vision, one of the main techniques for studying training robustness is via the use of backdoor attacks \cite{chou2023backdoor,yu2023backdoor,yuan2023you,jiang2023color,liu2020reflection} where the attacker injects stealthy backdoors into the victim model by poisoning a few samples in the training data. The attacker achieves this by attaching a trigger (i.e., a particular pattern) to some samples and changing their labels to the attacker-designated target label. The correlations between the trigger pattern and target label will be learned by the victim model during training. Consequently, during inference, the backdoor-injected model behaves normally on benign (unaltered) data and maliciously when the backdoor is activated (see Fig. \ref{fig:first_image}). The risk of such attacks has been recognized as one of the major areas of concern in autonomous navigation and driving \cite{jiang2020poisoning}.


To this end, we propose a novel adversarial backdoor attack designed to maliciously alter the performance of trajectory prediction models in AD systems. Our attack generates naturalistic (stealthy) poison samples through a novel disguising approach. Here, in an adversarial scheme, the adversary first generates triggers by introducing perturbations to the trajectory of the attacking vehicle and assigns a malicious future trajectory to it. Then, to make the trigger stealthy while preserving the attack's effect, using the proposed disguising method, the adversary conceals the generated trigger as a transformation of the road layout obtained by a bi-level optimization technique. The proposed attack can operate in a black-box manner and is model-agnostic, meaning that it does not require the surrogate model (the model used for generating poisoned samples) to have a similar architecture compared to the victim model. To the best of our knowledge this is the first adversarial backdoor attack designed for trajectory prediction models.

In the proposed attack, the trigger is the attacking vehicle (AtV)'s observation, which is a specific pattern designed by the attacker. Hence, the trigger is a rare yet feasible, i.e. realistic, trajectory pattern that can naturally be caused by the surrounding agents of the autonomous vehicle (AV) in real-world. In our method, the desired malicious outcome produced by the backdoor-injected model is achieved by intentionally inserting specific spurious correlations into the training set, exploiting the vulnerability of the model to learn those correlations. Therefore, the attack can be regarded as a method to discover the worst case predictions for potential spurious correlations in the training data as a means of determining the robustness of prediction models.


\noindent \textbf{Contributions:} 1) To the best of our knowledge, for the first time we study the vulnerability of trajectory prediction models from the viewpoints of data safety and security,  and robustness against potential spurious data correlations. For this, we propose a novel adversarial backdoor attack by data poisoning at training stage. Our model benefits from a novel bi-level optimization technique to disguise triggers as naturalistic and stealthy, hence, invisible to the victim; 2) To determine the effectiveness and noticeability of the attack, we propose modifications to the existing metrics and conduct extensive empirical studies using state-of-the-art trajectory prediction models on two AD benchmark datasets and highlight the impact of our proposed attack under various conditions; 3) Via performing ablation studies, we analyze the effectiveness of the attack under different constraints, and finally 4) we examine the capability of the existing defence mechanisms against the proposed backdoor attack.

\vspace{-2mm}

%% file: sec/2_related_work.tex
\section{Related Works}

\textbf{Trajectory Prediction. }The objective of trajectory prediction models in AD is to forecast the future coordinates of the road users for a given time horizon. There is a large body of work in this domain proposing approaches based on diverse architectural designs, including convolutional neural networks \cite{bansal2018chauffeurnet,chai2019multipath,salzmann2020trajectron++}, graph neural networks \cite{Rowe_2023_CVPR, Grimm_2023_ICRA,Pourkeshavarz_2023_ICCV,khandelwal2020if,zeng2021lanercnn}, and more recently, transformers \cite{Fang_2023_CVPR, Jiang_2023_CVPR, Zhou_2023_CVPR, Aydemir_2023_ICCV,huang2022multi,liu2021multimodal}. To make predictions, these models rely on encoding contextual information based on, for instance, rastersize images \cite{gao2020vectornet,zhao2021tnt,gu2021densetnt} or vectorized representations \cite{Aydemir_2023_ICCV, Fang_2023_CVPR,Grimm_2023_ICRA,Pourkeshavarz_2023_ICCV,gilles2022gohome} capturing the scenes and often dynamics alike. The latter representation is more dominant for encoding maps as it is more compact and efficient.  

\noindent \textit{\textbf{Robustness against attacks}. } Given the central role of trajectory prediction models in safe motion planning, their robustness to various adversarial attacks has been a major concern. Recently, a number of attacks have been proposed to study the vulnerability of these models. These attacks resort to carefully crafted perturbations applied to agents' trajectories \cite{zhang2022adversarial, cao2022advdo}, or scene layout \cite{bahari2022vehicle} and semantics \cite{Zheng_2023_WACV}. The attacks come in both types of white-box, where adversary has access to the victim model's parameters \cite{cao2022advdo, zhang2022adversarial} and black-box, in which the adversary does not have access \cite{bahari2022vehicle, zhang2022adversarial}. The proposed attack follows the second approach, and uses a different surrogate model than the victim to generate attacks. As we will show later, in our approach, the surrogate model does not necessarily need to have a similar architecture for the attacks to be effective.

The existing attack mechanisms only address the vulnerability of prediction models at inference time. However, the attacks can also occur at the training time by using approaches, such as backdoor attacks to poison the training data. Consequently, the victim model would behave maliciously when the backdoors are triggered. Such attacks are generally harder to detect and more difficult to reverse as they are encoded into the victim model. In this paper, we propose a novel backdoor attack for poisoning trajectory data. To the best of our knowledge, this is the first backdoor attack designed for trajectory prediction in AD.  

\label{related_backdoor}
\textbf{Backdoor Attack. }A backdoor attack is a deep learning training-time threat that assumes the attacker can modify the training data and procedure of a given model. The earliest works on backdoor attacks mainly focus on image classification tasks \cite{gu2019badnets} aiming to encourage malicious behaviours in the classifiers. However, due to the widespread use of DNNs in the industry, especially in safety-critical applications, the backdoor attacks have also received substantial attention in other fields of computer vision \cite{chou2023backdoor,yu2023backdoor,yuan2023you}. 

A category of backdoor attacks involves poisoning training data by injecting malicious samples with embedded triggers as backdoors. Models trained on such data behave normally on clean samples (without a trigger) but will exhibit a certain behavior on samples containing a trigger. To be effective, the injected trigger should be unnoticeable (does not impact the performance of the model on clean data) and stealthy (to appear realistic). In image classification, this objective is achieved by using imperceptible perturbations as backdoor triggers, consequently restricting the differences between the triggered and clean images in either pixel level \cite{chen2017targeted,li2020invisible,zhong2020backdoor} or latent space representation \cite{doan2021backdoor,ren2021simtrojan, zhao2022defeat}. Some works also rely on more explicit perturbations, for instance, by altering the color space \cite{jiang2023color} or adding artificial reflections \cite{liu2020reflection}. However, these methods sacrifice the attack's robustness and can be defeated using common preprocessing techniques. To evaluate a backdoor attack in classification tasks, clean accuracy (CA) and attack success rate (ASR) are common metrics of choice. As for the former, the backdoor-injected model's classification accuracy is measured against a clean test set and for the latter, the accuracy is measured against a poisoned test set. Thus, backdoor attacks with higher CA and ASR are considered successful. However, there is a trade-off between these two metrics, and it is often challenging to maintain high performance in both. In the proposed attack, we employ a novel bi-level optimization technique to first generate effective triggers and second to disguise them by performing dynamically feasible transformations to make them unnoticeable. Furthermore, since backdoor attacks have been studied only in classification tasks, we present modifications of the existing metrics, making them suitable for evaluating the proposed attack in the trajectory prediction domain.

 \begin{figure*}[t]
	\centering
	\includegraphics[width=0.9\textwidth]{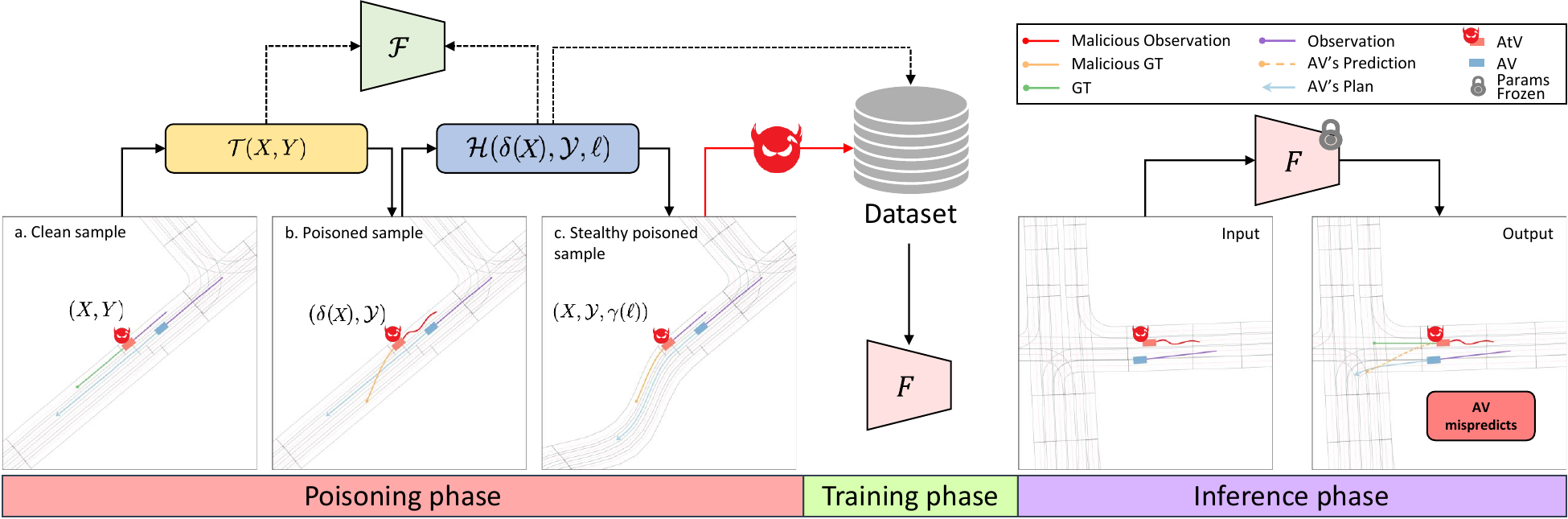}
 \vspace{-2mm}
	\caption {Overview of the proposed attack. At the poisoning phase, (a) given a benign sample, we select the closest vehicle to the AV with  observation $X$ and future trajectory $Y$ as the attacking vehicle (AtV). (b) Next, using the trigger generation method $\mathcal{T}(.)$, we obtain AtV's perturbed observation $\delta(X)$, malicious observation acting as the trigger, and malicious future trajectory $\mathcal{Y}$ to craft a poisoned sample. (c) Then, through the proposed backdoor disguising engine $\mathcal{H}(.)$ we apply transformations to the lanes $\gamma(\ell)$ to make the samples stealthy. We use a surrogate model $\mathcal{F}$ and have access to the dataset. The final $k$ stealthy poisoned samples are injected into the training data. In training phase, the victim trains a model on the poisoned dataset, resulting in a backdoor-injected model $F$. During inference, given an input sample including an AtV with a malicious observation, the backdoor-injected model mispredicts the AtV's future trajectory.}
	\label{fig:proposed_attack}
	\vspace{-4mm}
\end{figure*}

\vspace{-1mm}

%% file: sec/3_method.tex
\vspace{-1mm}
\section{Methodology}


\subsection{Preliminary}
\label{preliminary} \label{requirements}
\vspace{-1mm}
\textbf{Threat Model and Attack Requirements. }We follow the common threat model in the literature \cite{fowl2021preventing,geiping2021witches,zhao2021dataset} and define two parties, the \textit{attacker} and the \textit{victim}. The attacker can only manipulate the training data by injecting a small number of \textit{poisoned samples} and has no access to the victim model and its training process. The poisoned samples are made by implanting a \textit{trigger} (a specific pattern designed by the attacker) into the \textit{benign samples} (the original data samples) and changing the correct ground-truth to a malicious one. Then, the victim, unaware of the attack, trains a model on the \textit{poisoned data} resulting in a \textit{backdoor-injected model} that learns the association between the trigger and malicious ground-truth. Consequently, in inference time, the backdoor-injected model generates correct predictions given the benign samples (without the trigger) and malicious predictions given the samples altered by the trigger.

To be effective, a backdoor attack should be:

\noindent\uline{\textit{\textbf{Unnoticeable}}}. 
When evaluated on benign samples, the backdoor-injected model should perform similarly to the model trained on the original data. To validate this property, we measure the backdoor-injected model accuracy on the benign test set using clean accuracy (CA) metric.

\noindent \uline{\textit{\textbf{Effective}}}.
For the inputs altered by the trigger (poisoned samples) the backdoor-injected model should generate malicious predictions, i.e. it should behave the way the attacker desires. To verify this property, the backdoor-injected model is evaluated on the poisoned test set formed by adding the trigger to the benign test set and the performance is measured by attack success rate (ASR) metric.

\noindent\uline{\textit{\textbf{Stealthy}}}. The injected poisoned samples should not be recognized as abnormal samples by the victim. Therefore, the poisoned samples should 1) be perceived \textit{naturalistic}, i.e. have similar properties to the samples in the original data and 2) appear to have correct ground-truths, or \textit{clean label} as termed in the backdoor attack literature.

\textbf{Problem Formulation. }The prediction modules in AD systems forecast future trajectories of surrounding agents according to their past observed behavior. Specifically, at time step $t$, let the past trajectory of the $i$-th vehicle be a set of $2D$ coordinates in bird's eye view over some observation horizon $O$ time steps $X_i = \{(x_i, y_i)^{t-O+1}, \cdots,(x_i, y_i)^{t}\}$. Accordingly, the objective is to predict future trajectory $Y_i = \{(x_i, y_i)^{t+1}, \cdots,(x_i, y_i)^{t+T}\}$, where $T$ is the prediction horizon. The road information of the scene as an HD map represented in the vector space is also provided. Each scene is a matrix of stacked $2D$ coordinates consisting of all lane points in the $xy$ coordinate space where each row represents a point $(l_x, l_y)$ \cite{chang2019argoverse,bahari2022vehicle}. For simplicity, in the remainder of this paper, we refer to observations, future predictions, and lanes as $X$, $Y$, and $\ell$, respectively.

\vspace{-1mm}
\subsection{Attack Overview} \label{overview}
As illustrated in Fig. \ref{fig:proposed_attack}, our attack consists of three phases. During the poisoning phase, given a clean (benign) sample we first generate a trigger in the form of a trajectory of a vehicle, termed attacking vehicle (AtV). The trajectory is over the stochastic observation horizon with a malicious future path, such as turning towards the autonomous vehicle (AV). We refer to samples containing AtVs as poisoned. These samples are made stealthy using by utilizing a novel backdoor disguising method to generate \textit{naturalistic} trajectories with \textit{clean labels}.  In the training phase, the model is trained on the poisoned dataset resulting in a backdoor-injected model, which would behave maliciously when given a sample occupied by a trigger.

\vspace{-1mm}
\subsection{Trigger Generation} \label{trig-gen}
The trigger is a specific pattern designed by the attacker to craft a poisoned sample (Fig. \ref{fig:proposed_attack}.b) by adding a malicious behavior to a benign sample (Fig. \ref{fig:proposed_attack}.a). We design the trigger based on the trajectory of the AtV, selected as the closest vehicle to the AV, over the stochastic observation horizon. The trigger generation is therefore viewed as a perturbation $\delta(X)$, that is a minor change of the spatial coordinates $(\Delta x,\Delta y)$ of the AtV over the observation horizon.

To design the trigger, we use an adversarial scheme to find the perturbation that yields the desired outcome \cite{zhang2022adversarial,cao2022advdo}. Following the past works, we assume the attacker uses a surrogate model $\mathcal{F}$ which can be either the same as the victim model or a different one. We then define a set of constraints on the perturbation and an adversarial objective, e.g. a loss function, to find the perturbation that maximizes the attacker's objective. We impose physical constraints on the perturbations using a kinematic bicycle model \cite{cao2022advdo} to ensure that the altered trajectories are realistic. Then, we define the adversarial objective as follows:
\begin{equation}
\mathcal{L}=\mathcal{L}_{\mathrm{adv}}(Y, \hat{\mathcal{Y}})+
\alpha \mathcal{L}_{\mathrm{dyn}}\left(\delta(X)\right)
\end{equation}
where $\delta(X)$ stands for the perturbed trajectory and $\mathcal{L}_{\mathrm{dyn}}$ bounds the dynamic parameters by coefficient $\alpha$. $\mathcal{L}_{\mathrm{adv}}$ denotes the metrics for evaluating the prediction error. For this, we use average displacement error (ADE),  final displacement error (FDE) (error of the last predicted time step), and two additional metrics, namely average deviation towards the left and right side of the lateral direction \cite{zhang2022adversarial}.  

\vspace{-1.51mm}
\subsection{Backdoor Disguising} \label{Backdoor-dis}
The proposed backdoor disguising method aims to make the poisoned samples stealthy (Fig. \ref{fig:proposed_attack}.b-c), thus, they are not identifiable as abnormal. For this purpose, we disguise the generated triggers, AtV's altered trajectory, by transformations on the lanes $\gamma(\ell)$ under a condition $\mathcal{C}$ to create a clean label (i.e. future ground-truth that seems to be correct) for the poisoned sample. To achieve this goal, we define the following bi-level objective:
\begin{equation}
\begin{aligned}
\min _{\gamma \in \mathcal{C}} & \mathbb{E}_{(\smallx{}, \ell, \smally{}) \sim \mathcal{D}_k}[\mathcal{L}(\mathcal{F}  \boldsymbol{(} \delta(\mathbf{\smallx{}}), \mathbf{\ell}  ; \theta(\gamma) \boldsymbol{)}, \mathbf{\smallyy{}})] \\
\text { s.t. } \theta(\gamma) & \in \underset{\theta}{\arg \min } \sum_{(\smallx{}, \ell, \smally{}) \sim \mathcal{D}} \mathcal{L}(\mathcal{F}\boldsymbol{(} \mathbf{\smallx{}}, \gamma({\mathbf{\ell}}) ; \theta\boldsymbol{)}, \smally{})
\label{eq:bilevel}
\end{aligned}
\vspace{-1mm}
\end{equation}
where $X$ and $Y$ are AtV's observation and future trajectories before the alteration, and$\delta(X)$, $\mathcal{Y}$ stand for AtV's altered observation (trigger) and malicious future trajectories, respectively. $D$ and $\mathcal{D}_k$ denote the training and poisoned sets with sizes $m$ and $k$ $(k << m)$. We set $\mathcal{L}$ consistent with the objective $\mathcal{L}_{\mathrm{adv}}$ used in the trigger generation step.  

The condition $\mathcal{C}$ is defined as a constraint on the transformation to ensure that the transformed lanes make the AtV's label clean. Specifically, we check whether the AtV's malicious future trajectory $\mathcal{Y}$ is perceived as a correct ground-truth, e.g. no off-road, with the heading angle of trajectory aligned with the altered lane direction. 

For lane transformation function $\gamma(.)$, we define a general form of point transformation \cite{bahari2022vehicle} described below:\vspace{-1mm}
\begin{equation}
\tilde{l}=\left(l_x, l_y+f\left(l_x- r\right)\right)
\vspace{-1mm}
\end{equation}
where $\tilde{l}$ is the transformed point, $f$ is a differentiable single-variable transformation function, and $r$ is the reference point that defines the starting point of the transforming area. It should be noted that the transformation also applies to the non-malicious trajectories that are on the transformed lanes. 

To maintain the feasibility of the transformed trajectories, we determine whether the length of the trajectories is less than the maximum feasible displacement achievable in the given time horizon on the altered lanes \cite{halliday2013fundamentals,bahari2022vehicle} and clip the trajectories if their length exceeds to satisfy the check.

To circumvent the difficulty of bi-level optimization in Eq. \ref{eq:bilevel}, we approximate it using gradient alignment technique \cite{fowl2021preventing,souri2022sleeper,geiping2021witches} to modify the data to align the training gradient with the gradient of some desired objective. Contrary to other heuristics, e.g. partial unrolling of the computation graph, gradient alignment is a more stable way to solve a bi-level problem that entails training a network in an inner objective \cite{fowl2021preventing}. We define the adversarial objective as: 
\begin{equation}\vspace{-1mm}
\mathcal{L}_{\text{att}}=\mathbb{E}_{(\smallx{},\ell,\smally{}) \sim \mathcal{D}}\left[\mathcal{L}\left(\mathcal{F}\boldsymbol{(} \delta(\smallx{}) ; \theta\boldsymbol{)}, \smallyy{}\right)\right]
\label{eq:2}\vspace{-1mm}
\end{equation}
which is minimized when, given AtV's observed trajectory $\delta(\smallx{})$, the model mispredicts AtV's future as a malicious  trajectory $\mathcal{Y}$. For this, we perturb the training data by optimizing the following alignment objective:
\begin{equation}\vspace{-1mm}
\mathcal{A}=1-\frac{\nabla_\theta \mathcal{L}_{\text {train }} \cdot \nabla_\theta \mathcal{L}_{\text{att}}}{\left\|\nabla_\theta \mathcal{L}_{\text {train }}\right\| \cdot\left\|\nabla_\theta \mathcal{L}_{\text{att}}\right\|}
\label{eq:ga}\vspace{-1mm}
\end{equation}
where our goal is to find the transformation on the lanes $\gamma(\ell)$ that will make $\nabla_{\theta} \mathcal{L}_{\text{train}}$ to be aligned with $\nabla_{\theta} \mathcal{L}_{\text{att}}$. 
\begin{equation}\vspace{-2mm}
\nabla_\theta \mathcal{L}_{\text {train}}=\frac{1}{m} \sum_{i=1}^m \nabla_\theta \mathcal{L}\left(F\left(\smallx{}, \gamma(\ell) ; \theta\right), \smally{}_i\right)
\label{eq:nabla}\vspace{-2mm}
\end{equation}
Given the training gradient involving the nonzero transformations $\nabla_\theta \mathcal{L}_{\text {train}}$, we estimate the expectation in Eq. \ref{eq:2} by calculating the average adversarial loss as follows:
\begin{equation}
\nabla_\theta \mathcal{L}_{\text{att}}=\frac{1}{k}  \sum_{i=1}^k \nabla_\theta\left(\mathcal{L}\left(F\boldsymbol{(}\delta(\smallx{}), \ell ; \theta\boldsymbol{)}, \smallyy{}_i\right)\right).
\vspace{-2mm}
\end{equation}
We first optimize Eq. \ref{eq:ga} by fixing parameters $\theta$ to calculate $\mathcal{A}$ throughout the crafting process. The parameters are trained on the benign data and used to calculate the training $\nabla_\theta \mathcal{L}_{\text {train }}$ and adversary $\nabla_\theta \mathcal{L}_{\text {att }}$ gradients. We then optimize parameters $\theta$ using $s$ steps of signed Adam \cite{balles2018dissecting}. The complete method is summarized in Algorithm \ref{algo}.

\begin{algorithm}[t]
\SetKwInOut{Output}{Output}
\DontPrintSemicolon
\SetKwInOut{Input}{Input}
\SetAlgoLined
\SetKwInOut{Require}{Require}
\LinesNumbered 
 \Input{Training Data $D$, Poisoning budget $k$, Optimization steps $S$, Retraining factor $T$, Surrogate network $\mathcal{F}$}
 \Output{Poison perturbations} 
  {$\delta(\smallx{}), \smallyy{} \leftarrow \mathcal{T}(\smallx{},  \smally{})$}  \tcp{trigger generation} 
  {randomly initialize perturbations $\gamma(\ell)_{i=1}^k$} \leavevmode\\
   \For{ $r=1$ to $S$ }{ 

         $\text {Compute } \mathcal{A}\left(\gamma, \theta, \delta(\smallx{}), \smally{}, \smallyy{}\right) \text{using Eq.} \ref{eq:ga}$  \leavevmode\\
         
         $\text {Update } \gamma(\ell)_{i=1}^k \text { with a step of signed Adam }$ \leavevmode\\

           \uIf{$r \bmod \lfloor S /(T+1)\rfloor=0 \text { and } r \neq R $}{

            $\text{Poisoned training data }\mathcal{D} \leftarrow \left\{\left(\smallx{}_i, \gamma(\ell_i), \smallyy_i\right)\right\}_{i=1}^k \cup\left\{\left(\smallx{}_i, \ell_i , \smally{}_i\right)\right\}_{i=k+1}^n$\;
            
            $\text {Retrain network } \mathcal{F} \text { on } \mathcal{D}$\;
            
            $\text{Update network parameters } \mathcal{F}(. ;\theta$) \;}
 }
 \Return{$ \text{poison perturbations } \gamma(\ell)_{i=1}^k$}
   \caption{Proposed backdoor attack}
  \label{algo}
\end{algorithm} 
\setlength{\textfloatsep}{1pt}

Another aspect of the optimization is selection of benign samples for trigger injection. To achieve this, we resort to sample selection by gradient norm. More superficially,  we align the training gradient with our adversary objective and aim to select the samples that have larger gradients since such samples can be more potent poison instances. 

\newcolumntype{g}{>{\columncolor{MozhganGray}}c}
\begin{table*}[]
\centering
\caption{Overall attack results on different models evaluated on nuScenes and Argoverse. ADE and FDE metrics represent the performance of the clean models without any attacks, hence smaller values are better as indicated by $(\downarrow^*)$ . Benign/poison column shows the performance of the backdoor-injected models on the original validation set (no attacks) and the validation set with poisoned samples, respectively. For all metrics after attack, shown on highlighted gray area, higher values $(\uparrow)$ mean the attack was more successful.}
\label{tab:main_results}
\vspace{-2mm}
\resizebox{\textwidth}{!}{%
\begin{tabular}{l|l|l|cgggcggg}
\hline
\multirow{2}{*}{\textbf{Dataset}} & \multirow{2}{*}{\textbf{\makecell{Surrogate\\ model}}} & \multirow{2}{*}{\textbf{\makecell{Backdoor-\\injected model}}}  & \multicolumn{4}{c|}{\textbf{ADE}} & \multicolumn{4}{c}{\textbf{FDE}} \\
 &  &  & \multicolumn{1}{l}{{\textbf{original}$(\downarrow^*)$}} & \textbf{benign/poison} ($\uparrow$) & \textbf{tCA}$(\uparrow)$ & \multicolumn{1}{g|}{\textbf{tASR}$(\uparrow)$} & \multicolumn{1}{l}{\textbf{original}$(\downarrow^*)$} & \textbf{benign/poison} ($\uparrow$) & \textbf{tCA}$(\uparrow)$ & \textbf{tASR}$(\uparrow)$ \\ \hline
 \multirow{2}{*}{nuScenes} & PGP \cite{deo2022multimodal} & LaPred \cite{kim2021lapred} & 1.22 & 1.47/2.98 & \underline{89.13} & \multicolumn{1}{g|}{\textbf{76.98}} & 2.24 & 2.31/4.89 & \underline{86.32} & \textbf{93.12} \\
 & LaPred \cite{kim2021lapred} & PGP \cite{deo2022multimodal} & 0.94 & 1.06/2.67 & \textbf{91.01} & \multicolumn{1}{g|}{\underline{74.21}} & 1.55\ & 1.71/3.69 & \textbf{87.12} & \underline{83.45} \\ \hline \hline
 \multirow{12}{*}{Argoverse} & TNT \cite{zhao2021tnt} & \multirow{3}{*}{HiVT \cite{zhou2022hivt}} & \multirow{3}{*}{0.66} & 0.82/3.54 & \underline{94.86} & \multicolumn{1}{g|}{\underline{91.00}} & \multirow{3}{*}{0.96} & 1.10/5.36 & \underline{96.02} & 92.66 \\
 & MMTrans. \cite{liu2021multimodal}  & &  & 0.75/3.67 & \textbf{95.30} & \multicolumn{1}{g|}{\textbf{91.11}} &  & 1.04/5.21 &  \textbf{96.88} & \textbf{93.12} \\
 & LaneGCN \cite{liang2020learning} &  &   & 0.88/3.43 & 94.06 & \multicolumn{1}{g|}{90.88} &  & 1.21/5.48 & 95.39 & \underline{92.89}  \\ \cline{2-11} 
 & HiVT \cite{zhou2022hivt} & \multirow{3}{*}{TNT \cite{zhao2021tnt}}  & \multirow{3}{*}{0.95} & 2.09/2.83 & 73.16 & \multicolumn{1}{g|}{\underline{88.22}} & \multirow{3}{*}{1.73} & 3.02/4.36 & 75.72 & \underline{86.79} \\
  & MMTrans. \cite{liu2021multimodal} &  &  & 1.14/3.61 & \textbf{92.01}  & \multicolumn{1}{g|}{\textbf{90.74}} &  & 2.13/5.35 & \textbf{94.73} & \textbf{89.96} \\
  & LaneGCN \cite{liang2020learning} &  &  & 1.32/2.76 & \underline{89.46} & \multicolumn{1}{g|}{87.39} &  & 2.28/4.10 & \underline{90.33} & 85.29 \\ \cline{2-11} 
 & HiVT \cite{zhou2022hivt} & \multirow{3}{*}{MMTrans. \cite{liu2021multimodal}} & \multirow{3}{*}{0.70} & 1.39/3.16 & \textbf{84.21} & \multicolumn{1}{g|}{\underline{78.55}} & \multirow{3}{*}{1.08} & 2.31/4.40 & \textbf{85.67} & \textbf{83.44} \\
  & TNT \cite{zhao2021tnt} &  &  & 1.56/2.91 & 80.37 & \multicolumn{1}{g|}{77.89} &  & 2.61/4.06 & 83.01 & \underline{80.71} \\
  & LaneGCN \cite{liang2020learning} &  &  & 1.49/3.05 & \underline{82.40}  & \multicolumn{1}{g|}{\textbf{79.99}} &  & 2.48/4.14 & \underline{85.06} & 80.31 \\ \cline{2-11} 
 & HiVT \cite{zhou2022hivt} & \multirow{3}{*}{LaneGCN \cite{liang2020learning}} & \multirow{3}{*}{0.71} & 1.09/3.02 & \underline{87.00} & \multicolumn{1}{g|}{83.65} & \multirow{3}{*}{1.08} & 1.78/3.46 & 88.92 & 84.19 \\
  & TNT \cite{zhao2021tnt} &  &   & 1.12/2.92 & 86.46 & \multicolumn{1}{g|}{\textbf{89.24}} &  & 1.88/3.48 & \underline{89.65} & \textbf{86.99} \\
  & MMTrans. \cite{liu2021multimodal} &  &  & 1.01/3.48 & \textbf{88.02} & \multicolumn{1}{g|}{\underline{88.23}} &  & 1.59/3.61 & \textbf{89.79} &  \underline{86.36}\\ \hline 
\end{tabular}%
}
\vspace{-5mm}
\end{table*}

\vspace{-1mm}
\subsection{Evaluating Attacks} 
As discussed in  Sec. \ref{related_backdoor}, there are two widely used metrics, namely clean accuracy (CA) and attack success rate (ASR), for evaluating backdoor attacks. These metrics, however, are used for discrete classification tasks and are not directly applicable to trajectory prediction. Hence, we propose modifications too these metrics, referring to them as tASR and tCA where t stands for trajectory. For tCA, we compute ADE or FDE of the backdoor-injected model and clean model (the model trained with the original training set) on the benign validation data. We compare the errors for both models sample-wise, and if the degradation of a sample error is less than a threshold $th_1$, we consider that instance a correct prediction, otherwise incorrect. tCA is then calculated as the ratio of the correct predictions over all predictions. Similarly, for tASR, we compute the errors for both backdoor-injected and clean models on the poisoned validation set and compare them. If degradation of the error on a sample is more than a threshold $th_2$, then the attack is successful, otherwise it is not. tASR is then calculated as the ratio of successful attacks over all attacks. 

The way the thresholds are set is important as they should correspond to real driving conditions. Given that urban lanes are approximately $3.7m$ wide and cars are on average  $1.7m$ wide, a $1m$ deviation is an acceptable deviation for a car not to drive off-road or into another lane when normally driving in the center of a lane \cite{hong2020effectiveness}. Therefore, we set $th_2=1m$. For clean accuracy, however, we consider a stricter threshold of  $th_1=0.5m$ to increase the sensitivity of this metric to slight deviations as it corresponds to the detectability of the attacks not their success rate.

\vspace{-2mm}

%% file: sec/4_experiments.tex
\section{Experiments}
For evaluation, we seek to answer the following questions: 
$\textbf{Q1:}$ Does the proposed attack remain unnoticeable despite its effectiveness? $\textbf{Q2:}$ For a successful attack, how many poisoned samples should be injected into the training dataset? $\textbf{Q3:}$ How many AtVs are required to successfully launch the attack? $\textbf{Q4:}$ When making poisoned samples, does the choice of benign samples affect the attack success rate? $\textbf{Q5:}$ Is the proposed attack still effective with partial access to the training data? $\textbf{Q6:}$ Does the proposed attack work across different representation encoding?   $\textbf{Q7:}$ Are the existing defences effective against the proposed attack?


\noindent\textbf{Datasets}
We use two widely-used large-scale autonomous driving datasets, namely \textit{nuScenes} \cite{chang2019argoverse} and \textit{Argoverse} \cite{caesar2020nuscenes}, which are catered for trajectory prediction task. nuScenes contains 1K driving scenes, each of which are $20s$ long and annotated at $2 \mathrm{~Hz}$. For this benchmark the task is to predict $6s$ future trajectory given $2s$ observation. The sequences extracted from the driving scene are split train, validation and test sets with 32K, 8.6K and 9K instances in each respectively. Argoverse consists of over $30 \mathrm{~K}$ driving scenarios, each sampled at $10 \mathrm{~Hz}$. Here, the task is to predict $3s$ future trajectory of a road agent given $2s$ observations. The sequences in this data are split into train, val, and test sets with 206K, 39K, and 78K sequences in each respectively.

\noindent\textbf{Models.} On each dataset, we evaluate the state-of-the-art (SOTA) models from the corresponding benchmark leaderboards. For nuScenes, we choose \textit{PGP} \cite{deo2022multimodal} and \textit{LaPred} \cite{kim2021lapred} and for Argoverse, we select \textit{HiVT} \cite{zhou2022hivt}, \textit{TNT} \cite{zhao2021tnt}, \textit{MMTransformer} \cite{liu2021multimodal}, and \textit{LaneGCN} \cite{liang2020learning}, which are representative of different approaches. We use official code released for all modes, with the exception of TNT\footnote{The implementation used is from \url{https://github.com/Henry1iu/TNT-Trajectory-Prediction}}.

\textbf{Implementation.} We set  $R=4$ for solving the bi-level optimization. For the transformation function, we use $f = \gamma_1\left(1-\cos \left(2 \pi \gamma_2 s_x\right)\right)$ for $l_x\geq0$ and zero for the rest of the points, where $\gamma_1$ and $\gamma_2$ determine the turn curvature and the sharpness of the turns. We empirically choose $\gamma_1 = 5.75$ and $\gamma_2 = 0.015$ for best results. We set the number of steps in the alignment process as $S=250$. 

\begin{table*}[h!]
\centering
\caption{tASR and tCA metric values with different numbers of AtVs (q) as the trigger for different backdoor-injected models on the Argoverse dataset. $(\downarrow)$ and $(\uparrow)$ show lower and higher values are better.}
\vspace{-2mm}
\label{tab:my-table}
\resizebox{0.7\textwidth}{!}{%
\begin{tabular}{l|cccccccc}
\hline
 & \multicolumn{8}{c}{\textbf{Backdoor-injected model}} \\
 & \multicolumn{2}{c}{\textbf{HiVT \cite{zhou2022hivt}}} & \multicolumn{2}{c}{\textbf{TNT \cite{zhao2021tnt}}} & \multicolumn{2}{c}{\textbf{MMTrans. \cite{liu2021multimodal}}} & \multicolumn{2}{c}{\textbf{LaneGCN} \cite{liang2020learning}} \\ \cline{2-9} 
\textbf{\#AtV / metric} & \multicolumn{1}{c}{\textbf{tCA}$(\uparrow)$} & \multicolumn{1}{c|}{\textbf{tASR}$(\uparrow)$} & \multicolumn{1}{c}{\textbf{tCA}$(\uparrow)$} & \multicolumn{1}{c|}{\textbf{tASR}$(\uparrow)$} & \multicolumn{1}{c}{\textbf{tCA}$(\uparrow)$} & \multicolumn{1}{c|}{\textbf{tASR}$(\uparrow)$} & \multicolumn{1}{c}{\textbf{tCA}$(\uparrow)$} & \multicolumn{1}{c}{\textbf{tASR}$(\uparrow)$} \\ \hline
q = 1 & 95.30 & \multicolumn{1}{c|}{91.11} & 92.01 & \multicolumn{1}{c|}{90.74} & 82.40 & \multicolumn{1}{c|}{79.99} & 86.46 & 89.24 \\
q = 2 & 92.03 & \multicolumn{1}{c|}{94.29} & 88.32 & \multicolumn{1}{c|}{92.65} & 79.36 & \multicolumn{1}{c|}{80.07} & 83.66 & 90.37 \\
q = 3 & 86.11 & \multicolumn{1}{c|}{95.47} & 86.21 & \multicolumn{1}{c|}{95.78} & 68.93 & \multicolumn{1}{c|}{84.63} & 75.47 & 91.58 \\
q = 4 & 71.34 & \multicolumn{1}{c|}{98.02} & 73.12 & \multicolumn{1}{c|}{96.22} & 60.16 & \multicolumn{1}{c|}{87.27} & 69.20 & 93.23 \\ \hline
\end{tabular}%
}
\vspace{-3mm}
\end{table*}

 \begin{figure*}[t!]
	\centering
 \resizebox{0.9\textwidth}{!}{
\subfloat{\label{figur:1}\includegraphics[width=0.232 \textwidth]{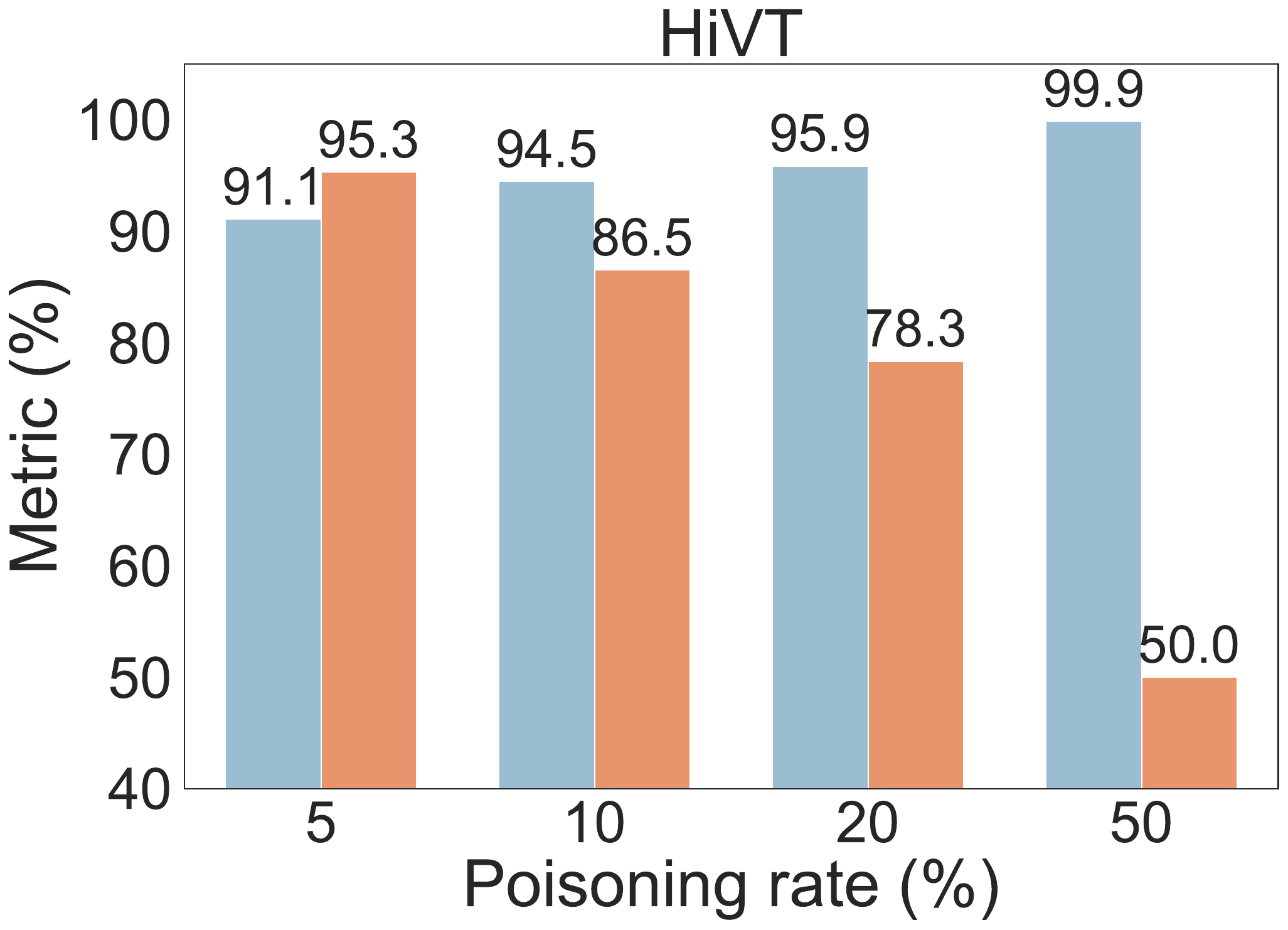}} 
\subfloat{\label{figur:2}\includegraphics[width=0.22 \textwidth]{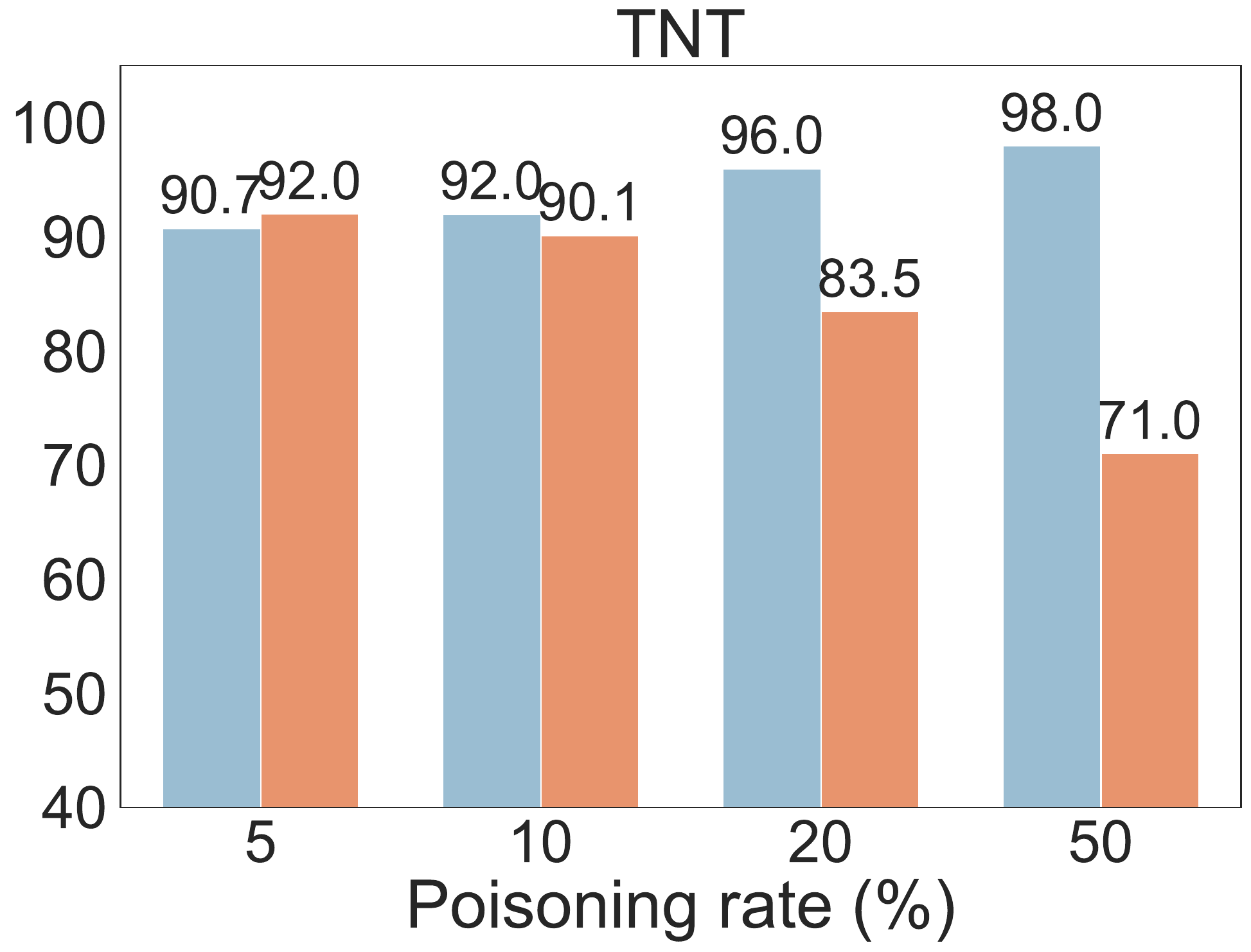}} 
\subfloat{\label{figur:3}\includegraphics[width=0.22 \textwidth]{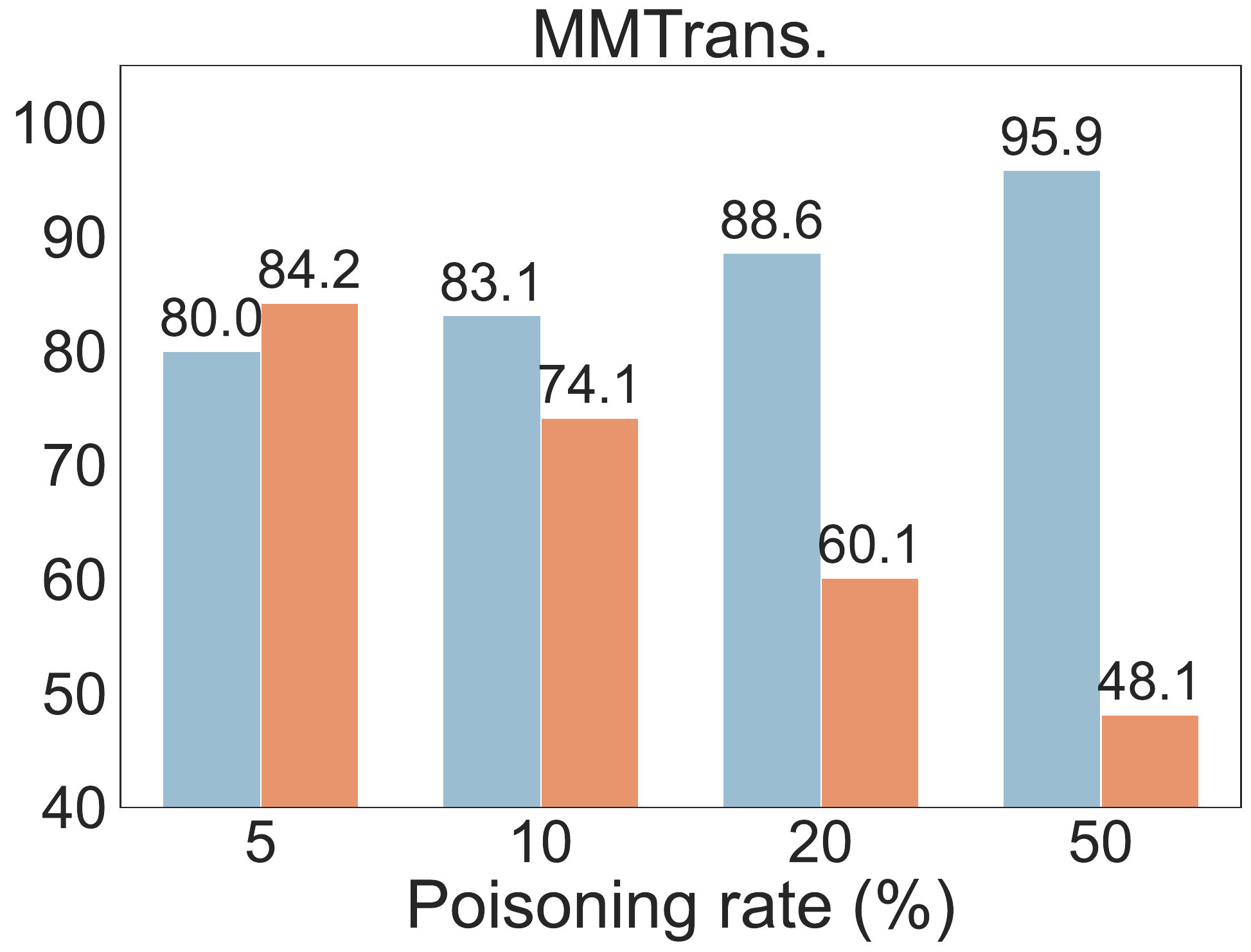}} 
\subfloat{\label{figur:4}\includegraphics[width=0.29\textwidth]{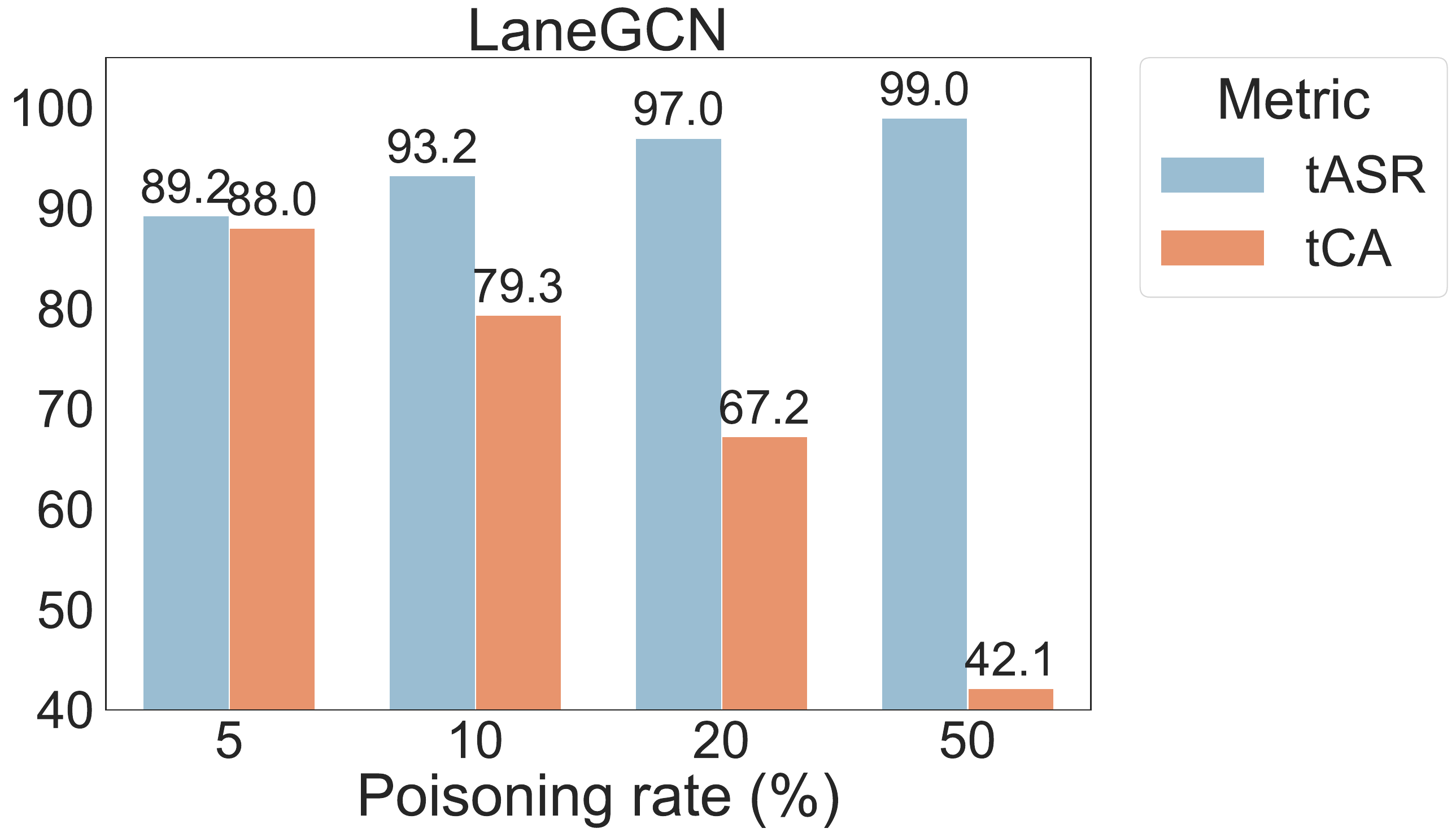}} }
\vspace{-2mm}
\caption{The effectiveness of proposed attack with varying poison rates for different backdoor-injected models on Argoverse .}
\label{fig:poison_budget}
\vspace{-4mm}
\end{figure*}

\begin{table*}[th]
\centering
\caption{tASR metric values with different selection mechanisms for different backdoor-injected models and varying poisoning rates on the Argoverse dataset. -G and -R stand for gradient and random selection, respectively. $(\uparrow)$ shows the attack is more successful.}
\vspace{-2mm}
\label{tab:selection}
\resizebox{0.85\textwidth}{!}{%
\begin{tabular}{l|cccccccc}
\hline
 & \multicolumn{8}{c}{\textbf{Backdoor-injected model}} \\
 & \multicolumn{2}{c}{\textbf{HiVT} \cite{zhou2022hivt}} & \multicolumn{2}{c}{\textbf{TNT} \cite{zhao2021tnt}} & \multicolumn{2}{c}{\textbf{MMTrans.} \cite{liu2021multimodal}} & \multicolumn{2}{c}{\textbf{LaneGCN} \cite{liang2020learning}} \\ \cline{2-9} 
\textbf{Pr / metric} & \multicolumn{1}{c}{\textbf{tASR-R}$(\uparrow)$} & \multicolumn{1}{c|}{\textbf{tASR-G}$(\uparrow)$} & \multicolumn{1}{c}{\textbf{tASR-R}$(\uparrow)$} & \multicolumn{1}{c|}{\textbf{tASR-G}$(\uparrow)$} & \multicolumn{1}{c}{\textbf{tASR-R}$(\uparrow)$} & \multicolumn{1}{c|}{\textbf{tASR-G}$(\uparrow)$} & \multicolumn{1}{c}{\textbf{tASR-R}$(\uparrow)$} & \multicolumn{1}{c}{\textbf{tASR-G}$(\uparrow)$} \\ \hline
Pr = 5 (\%)& 86.12 & \multicolumn{1}{c|}{91.11} & 83.67 & \multicolumn{1}{c|}{90.74} & 70.23 & \multicolumn{1}{c|}{79.99} & 70.69 & 89.24 \\
Pr = 10 (\%) & 90.39 & \multicolumn{1}{c|}{94.29} & 88.83 & \multicolumn{1}{c|}{92.65} & 73.43 & \multicolumn{1}{c|}{80.07} & 83.66 & 90.37 \\
Pr = 20 (\%)& 92.67 & \multicolumn{1}{c|}{95.47} & 92.00 & \multicolumn{1}{c|}{95.78} & 79.99 & \multicolumn{1}{c|}{84.63} & 86.58 & 91.58 \\
Pr = 50 (\%)& 95.89 & \multicolumn{1}{c|}{98.02} & 94.16 & \multicolumn{1}{c|}{96.22} & 81.32 & \multicolumn{1}{c|}{87.27} & 89.69 & 93.23 \\ \hline
\end{tabular}%
}
\vspace{-5mm}
\end{table*}

\vspace{-1mm}
\subsection{Black-box Backdoor Attack} \label{results}
We evaluate the proposed attack in a black-box fashion, meaning that we impose a restrictive condition and only allow the attacker to have access to the training dataset without any knowledge of the training model. Thus, in each experiment, a different surrogate model other than the backdoor-injected model is chosen. The experimental results are reported in Table \ref{tab:main_results}.

\noindent\textbf{Unnoticeable and effective.} The first glance at the results reveals the effectiveness of the proposed attack as the performance of all models on both datasets has significantly degraded on both benign and poison validation sets. The higher clean accuracy (tCA) values, especially for top performing models, such as PGP ($91\%$) on nuScenes and HiVT ($95\%$) on Argoverse, indicate that our attack is unnoticeable as these models learned the poison samples during the training phase without any major impact on their performance during validation on clean data. However, once exposed to poison samples, the models are significantly impacted, as indicated by high attack success ratios (tASRs), in the case of PGP and HiVT by more than $83\%$ and $93\%$ respectively. High tASR values show how effective the proposed attack is in forcing the models to generate malicious behavior at inference time. Overall, large values of tCA and tASR metrics across all models indicate that the proposed attack is unnoticeable and yet effective in altering the performance of the victim models even without having access to their architecture and parameters.

\noindent\textbf{Accuracy vs robustness.} In addition to measuring how unnoticeable the attacks are, tCA shows the robustness of the models to poisoned training data when evaluated under normal conditions. Hence, as shown in Table \ref{tab:main_results}, there is a correlation between accuracy and tCA values for top performing models. For instance, best performing models, PGP and HiVT on both datasets with the highest accuracy ($0.94/1.55$ and $0.66/0.96$) also have the highest overall tCA values ($91\%/87\%$ and $95\%/97\%$). However, higher accuracy does not necessarily translate to higher robustness. For example, TNT with the worst overall accuracy ($0.95/1.73$) has a better tCA ($92\%)$ compared to MMTransformer with higher accuracy ($0.70/1.08)$) but lower tCA ($80\%$). 

Similarly for tASR, for instance, the most accurate model on Argoverse,  HiVT, is at the same time the most vulnerable model to the proposed attack reaching the peak value of $93\%$. In terms of robustness in training vs inference, TNT with the second highest tCA value on Argoverse, is also the second worst model in terms of tASR, as this model is impacted significantly more by the proposed attack compared to MMTransformer and LaneGCN.  

\noindent\textbf{Choice of the surrogate model.} Last but not least, as shown in Table \ref{tab:main_results}, regardless of the choice of surrogate model, the proposed attack is very successful. This is highlighted in small fluctuations of tCA and tASR values for each backdoor-injected model with different surrogates. There are, however, exceptions as well. For instance, tCA value of TNT when trained on the poison data with HiVT as surrogate is significantly lower compared to other surrogate models. This can be due to the properties of the generated samples by HiVT that are not easily learnable for TNT.

\vspace{-1mm}
\subsection{Ablation Study}

\begin{table}[]
\centering
\caption{An ablation study of the attacker's partial access to the training dataset. $(\uparrow)$ shows the attack is more successful.}
\vspace{-2mm}
\label{tab:partia-acc}
\resizebox{\columnwidth}{!}{%
\begin{tabular}{l|cccc}
\hline
 & \multicolumn{4}{c}{\textbf{Backdoor-injected model}} \\
 & \multicolumn{2}{c|}{\textbf{HiVT}\cite{zhou2022hivt}} & \multicolumn{2}{c}{\textbf{MMTrans}\cite{liu2021multimodal}} \\ \cline{2-5} 
\textbf{\textbf{d / metric}} & \textbf{tCA}$(\uparrow)$ & \multicolumn{1}{c|}{\textbf{tASR}$(\uparrow)$} & \textbf{tCA}$(\uparrow)$ & \textbf{tASR}$(\uparrow)$ \\ \hline
d = 100 (\%) & \textbf{95.30}  & \multicolumn{1}{c|}{\textbf{91.11}} & \textbf{84.21} & \textbf{78.55} \\
d = 80 (\%) & 93.63 \leavevmode\color{red}(-1.67) & \multicolumn{1}{c|}{90.88 \leavevmode\color{red}(-0.23)} & 78.16 \leavevmode\color{red}(-6.05) & 79.69 \leavevmode\color{red}(+1.14) \\
d = 50 (\%) & 83.28 \leavevmode\color{red}(-12.02) & \multicolumn{1}{c|}{86.13 \leavevmode\color{red}(-4.98)} & 70.61 \leavevmode\color{red}(-13.60) & 72.37 \leavevmode\color{red}(-6.18) \\ \hline
\end{tabular}%
}
\vspace{-3mm}
\end{table}


\textbf{Number of AtVs. }
Thus far we showed the effectiveness of the proposed attack with triggers using only a single attacking vehicle (AtV). Here, we experiment with varying numbers of AtVs in poisoned samples. More specifically, we set the number of AtV as $q = \{1, 2, 3, 4\}$.In each, we designate the closest $q$ vehicles to the AV as AtVs.

As shown in Table. \ref{tab:my-table}, in general, higher the number of AtVs, the more impactful the attack is as it is evident in the rise of tASR  and fall of tCA values. This is because the larger number of AtVs increases the likelihood that the model learns the association between the trigger and malicious label. Hence, the model becomes more vulnerable to the attack. Once again, depending on the victim models, the impact of the attack may vary. Overall, HiVT has the highest drop ($24\%$) in tCA for $q=4$ and the highest gain in tASR, similar to MMTransformer, by approx. $7\%$. Among all models, with respect to both metrics, LanceGCN is generally least impacted by increasing the number of AtVs.

\textbf{Poison budget. }
To study the utility and specificity of the proposed attack, we vary the poisoning rate (budget) $Pr$. i.e. the fraction of modified training samples to all samples. The results for different poisoning rates,  $Pr=\{5\%, 10\%, 20\%, 50\% \}$ are illustrated in  Fig. \ref{fig:poison_budget}. For each model in Table \ref{tab:main_results}, we select the surrogate variation that resulted in the highest tASR value for $Pr=5$. As we can see the proposed attack is very efficient and can succeed with as small as $5\%$ poison budget. As expected, poisoning rate has direct and inverse relationship with tASR and tCA respectively. The higher the poisoning rate is, the higher tASR and the lower the tCA values are. The inclination of change, however, is different for the models. For instance, TNT has the lowest drop in tCA, with only $21\%$, reaching the top spot of $71\%$ on $50\%$ poison budget. LaneGCN, on the other hand, has the lowest tCA for the same budget at $42.1\%$ with the drop date of up to $46\%$. 

\textbf{Effect of sample selection. }
As mentioned in Sec. \ref{Backdoor-dis}, samples with larger gradients are selected to craft poison samples. To verify this approach, we conduct an experiment using a random selection mechanism with varying poisoning rates. As shown in Table \ref{tab:selection} using the proposed gradient-based selection approach is significantly more effective, especially for smaller poison budgets where an increase of up to $19\%$ is achieved. The reduction in gap between different selection mechanisms for larger budgets is expected as the likelihood of larger gradient samples being selected in the random procedure increases. 

\vspace{-2mm}
\textbf{Full vs. partial access. }
In previous experiments, the assumption was that the attacker has full access to the training data. Here, we conducted an experiment limiting the attacker's access to only a part of the training data. We randomly select $d\%$ of the training dataset and launch the proposed attack on two models, namely HiVT \cite{zhou2022hivt} and MMTrans \cite{liu2021multimodal} with the highest and the lowest robustness against the attacks on Argoverse according to Table \ref{tab:main_results}. Here, each model acts as the surrogate to design the attack against the other model. Based on the results in Table \ref{tab:partia-acc}, as expected, the effectiveness of the attack is lowered as the access to the training data is reduced. However, the ratio of attack's impact degradation is significantly lowere compared to the ratio of limiting data access. In the case of $d=80\%$ there is only a minor fluctuation in both models' tASR ($\approx 1\%$) and tCA ($\approx 7\%$) . When reducing the access to only $50\%$, the drop in tCA and tASR of both models do not exceed $17\%$ and $8\%$, respectively. This shows that the attack is still effective even with partial access to the data. Note that the higher drop ratio in tCA compared to tASR is expected as the attacks are optimized with respect to effectiveness rather than being unnoticeable.

\subsection{Cross-representation Backdoor Attack}
\begin{figure}[t]
	\centering
	\includegraphics[width=0.79\columnwidth]{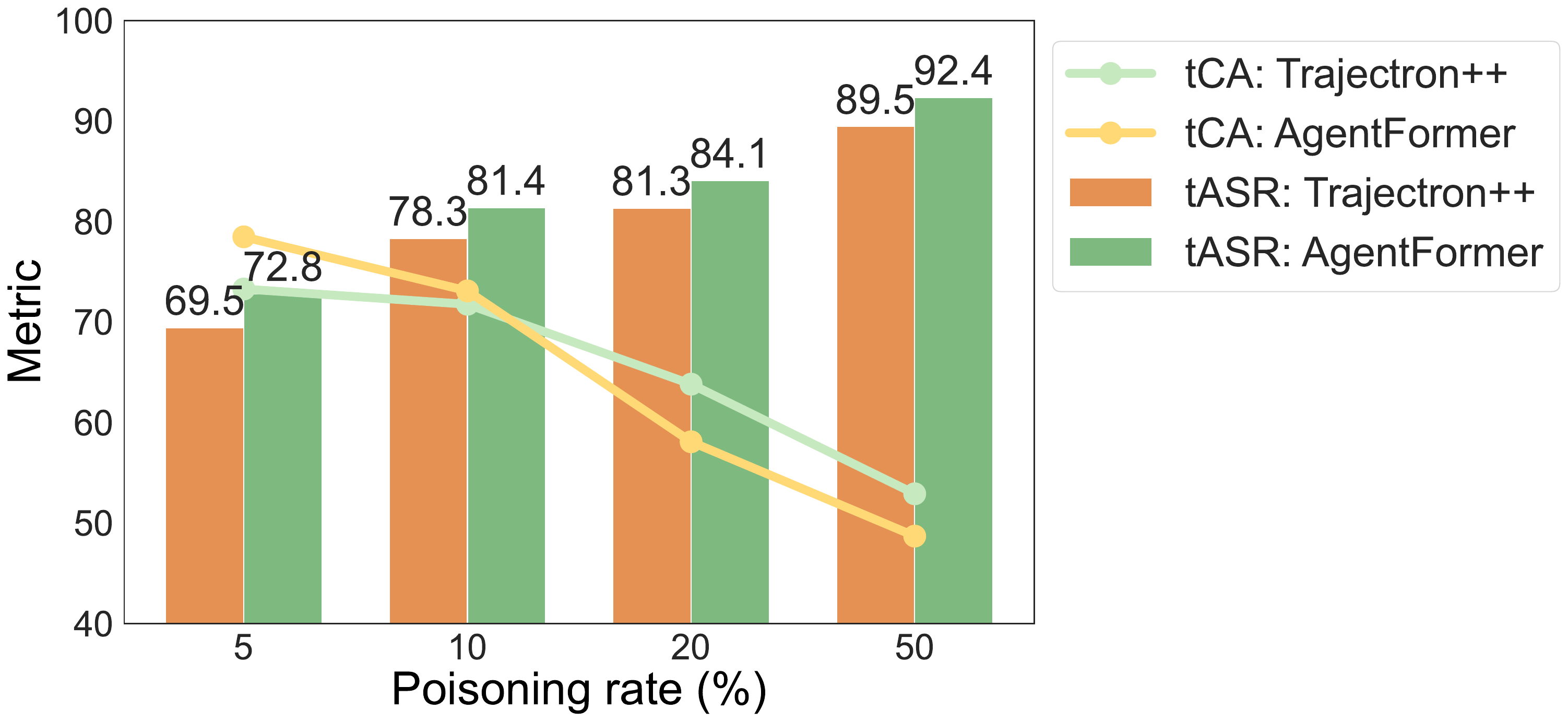}
 	\vspace{-2mm}
	\caption {The attack with varying poison rates on models with rasterized-based map encoder on nuScenes.}
	\label{fig:traj_agent}
	\vspace{-3mm}
\end{figure}

In our experiments thus far, despite architectural differences, the surrogate and victim models shared similar encoding mechanisms based on the vector map representation. Here, our goal is to determine whether attacks designed using a model with vector-based map encoder can be effective against the models consist of rasterized map encoders. To this end, we use PGP \cite{deo2022multimodal} as surrogate and two state-of-the-art models with rasterized map representation, namely  Trajectron++ \cite{salzmann2020trajectron++} and AgentFormer \cite{yuan2021agentformer}, as victims. As demonstrated in Fig. \ref{fig:traj_agent}, the proposed attack is still as effective as before. In fact, we can observe similar trends compared to vectorized surrogate-victim pairs. On $Pr$ equal to $5\%$ and $10\%$, we can observe high tCA and tASR values, which point to the efficiency and effectiveness of the proposed attack while staying inconspicuous. As before, by increasing the poison budget, there is an increase in tASR value and decline in tCA on both victim models.

\begin{table}[]
\centering
\caption{The tCA and tASR metrics with (w) and without (w/o) applying the defence. $(\uparrow)$ shows the attack is more successful.}
 \vspace{-2mm}
\label{tab:defence}
\resizebox{\columnwidth}{!}{%
\begin{tabular}{l|cc|cc}
\hline
 & \multicolumn{2}{c|}{\textbf{w/o defence}} & \multicolumn{2}{c}{\textbf{w defence}} \\
\textbf{Backdoor-injected model} & \textbf{tCA}$(\uparrow)$ & \textbf{tASR}$(\uparrow)$ & \textbf{tCA}$(\uparrow)$ & \textbf{tASR}$(\uparrow)$ \\ \hline
HiVT \cite{zhou2022hivt} & 95.30 & 91.11 & 93.33 \leavevmode\color{red}(-1.97) & 88.52 \leavevmode\color{red}(-2.59) \\
MMTrans. \cite{liu2021multimodal} & 84.21 & 78.55 & 81.62 \leavevmode\color{red}(-2.59) & 73.98 \leavevmode\color{red}(-4.57) \\ \hline
\end{tabular}%
}\vspace{3mm}
\end{table}

\vspace{-1.5mm}
\subsection{Defence and Mitigation}
Since the proposed attack is based on data poisoning in a black box setting, the defence mechanism should be deployed in the training time to detect poisoned samples before being fed in the victim model. Due to the stealthiness of the proposed attacks achieved by our backdoor disguising approach, the triggers (the AtV's malicious observation) used to induce mispredictions during inference time are not directly observable in the training dataset. This means that the existing preprocessing trajectory mechanisms \cite{zhang2022adversarial, cao2022advdo} would not be effective to mitigate the proposed attack.

Since poison samples are rare in the training data, detection-based defences using gradient shaping methods, which are effective against gradient alignment based attacks can be used \cite{hong2020effectiveness} . Following \cite{hong2020effectiveness}, during the training phase, the gradients of the weights that are perceived abnormal are clipped and perturbed by some adding noise to them in order to mitigate the effect of poisoned samples. We experimented using HiVT \cite{zhou2022hivt} and MMTransformer \cite{liu2021multimodal} models on Argoverse with the clipping and noise values from \cite{hong2020effectiveness} and report the best results with the highest attack mitigation impact in Table \ref{tab:defence}. As the findings suggest, although the defence is effective, the improvements are marginal, up to $3\%$ in tCA and $6\%$ in ASR. As a result, the attack stays very effective, maintaining over $70\%$ tASR on both models. The reason for this is that even though the injected transformations are rare, they are realistic thanks to our disguising method based on dynamically feasible constraints. 

\vspace{-2mm}

%% file: sec/5_conclusion.tex
\section{Conclusion}
We proposed a novel adversarial backdoor attack as a means of studying the vulnerability of trajectory prediction models in security-critical systems, such as autonomous driving. Our method is based on a novel bi-objective optimization process that generates attack triggers and effectively disguises them via realistic transformations. We conducted extensive empirical evaluations on state-of-the-art trajectory prediction models on common benchmark datasets and showed that our attack is not noticeable and significantly effective to force victim models to generate malicious predictions. Furthermore, we conducted ablative studies highlighting the effectiveness of the proposed attacks under constrained conditions and also showed that the existing defence mechanisms are not very effective in mitigating the impact of our attacks. Our work highlighted the potential danger of backdoor attacks in autonomous driving and the necessity of designing more robust algorithms and defence mechanisms to detect and mitigate the effect of such attacks.